\documentclass[10pt,twocolumn,letterpaper]{article}

\usepackage{iccv}
\usepackage{times}
\usepackage{amsmath}
\usepackage{amssymb}
\usepackage[pagebackref=true,breaklinks=true,letterpaper=true,colorlinks,bookmarks=false]{hyperref}
\usepackage{booktabs}
\usepackage{graphicx,multirow}
\usepackage{bm}
\usepackage{adjustbox}
\iccvfinalcopy 

\def\iccvPaperID{11059} 

\ificcvfinal\fi

\begin{document}

\title{Remote Sensing Change Detection with Transformers Trained from Scratch}

\author{Mubashir Noman\textsuperscript{1} Mustansar Fiaz\textsuperscript{1}   Hisham Cholakkal\textsuperscript{1} Sanath Narayan\textsuperscript{2}  \\  Rao Muhammad Anwer\textsuperscript{1,3} Salman Khan\textsuperscript{1,4} Fahad Shahbaz Khan\textsuperscript{1,5} \hspace{.1cm} 
\\
\textsuperscript{1}Mohamed bin Zayed University of AI, UAE \hspace{.1cm} \textsuperscript{2}Technology Innovation Institute, UAE
\hspace{.1cm}  \\ \textsuperscript{3}Aalto University, Finland  \hspace{.1cm} \textsuperscript{4}Australian National University \textsuperscript{5}Link{\"o}ping University, Sweden\\
}

\maketitle
\ificcvfinal\thispagestyle{empty}\fi

\begin{abstract}
Current transformer-based change detection (CD) approaches either employ a pre-trained model trained on large-scale image classification ImageNet dataset or rely on first pre-training on another CD dataset and then fine-tuning on the target benchmark. This current strategy is driven by the fact that transformers typically require a large amount of training data to learn inductive biases, which is insufficient in standard CD datasets due to their small size.
We develop an end-to-end CD approach with transformers that is trained from scratch and yet achieves state-of-the-art performance on four benchmarks. Instead of using conventional self-attention that struggles to capture inductive biases when trained from scratch, our architecture utilizes a shuffled sparse-attention operation that focuses on selected sparse informative regions to capture the inherent characteristics of the CD data. Moreover, we introduce a change-enhanced feature fusion (CEFF) module to fuse the features from input image pairs by performing a per-channel  re-weighting. Our CEFF module  aids in enhancing the relevant semantic changes while suppressing the noisy ones. Extensive experiments on four CD datasets reveal the merits of the proposed contributions, achieving gains as high as 14.27\% in intersection-over-union (IoU) score, compared to the best published results in the literature. Code is available at \url{https://github.com/mustansarfiaz/ScratchFormer}.

\end{abstract}

\begin{figure*}[t!]
\centering
    \includegraphics[width=0.88\textwidth]{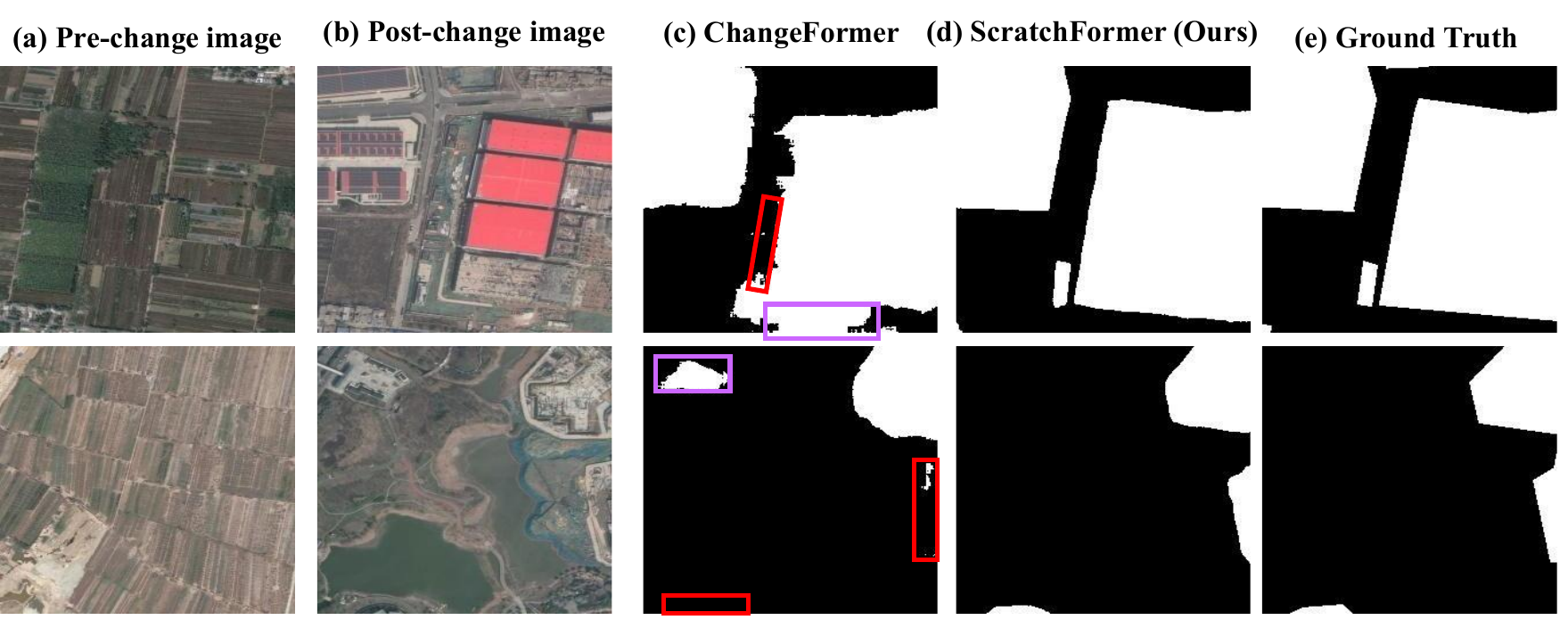} 
  \caption{Change detection (CD) performance comparison of \textbf{(d)} our approach (ScratchFormer) with \textbf{(c)} the recent ChangeFormer ~\cite{bandara2022transformer} on examples from DSIFN-CD dataset. Here, \textbf{(a)} the pre-change image and \textbf{(b)} post-change image are shown along with \textbf{(e)} the ground-truth. We show the false positives and false negatives in the purple and red color, respectively. In both rows, the ChangeFormer incorrectly detects a change region (purple box). Similarly, the ChangeFormer fails to detect the change occurring between the pre- and post-images (red box in both rows), as indicated in the ground-truth. Our ScratchFormer achieves improved change detection performance by reducing both false positive and negatives. Best viewed zoomed in. Additional results are in Fig. \ref{fig:comparison_with_SOTA_on_dsifn}, Fig. \ref{fig:comparison_with_SOTA_on_levir} and the supplementary.
  }
  \label{fig:baseline_comparison}
\end{figure*}

\section{Introduction}
\label{sec:intro}
Change detection (CD) is a fundamental remote sensing research problem that strives to identify all relevant changes between  co-registered satellite images acquired at distinct timestamps.
Here, the objective is to detect relevant semantic changes in man-made facilities such as, buildings and other constructions while ignoring noisy changes such as shadows and all types of  seasonal and environmental variations. CD plays a crucial role in various remote sensing applications including, disaster management \cite{kucharczyk2021remote}, urban planning \cite{yin2021integrating}, forestry and ecosystem monitoring \cite{fonseca2021pattern, wen2021change}. Generally, CD approaches relying on convolutional neural networks (CNN)  has shown promising results  by utilizing explicit mechanisms such as dilated convolutions, channel and spatial attentions \cite{zhang2018triplet, chen2020dasnet, zhang2020deeply, daudt2018fully, chen2020spatial}. However, these CNN-based approaches typically struggle to capture  long-range dependencies between different image regions, hampering the change detection performance.

Recently, transformer-based CD methods~\cite{bandara2022transformer, li2022transunetcd, song2022mstdsnet, wang2022network, ke2022hybrid, chen2021remote} have shown competitive performance on various CD datasets by capturing long-range dependencies between uniformly sampled dense patches through self-attention \cite{vaswani2017attention}. 
Although achieving superior CD performance, state-of-the-art transformer-based methods~\cite{bandara2022transformer} 
generally require \textit{pre-training} based weight initialization for optimal convergence. The pre-training step in existing transformer-based CD methods either involve another CD dataset ~\cite{bandara2022transformer} or an ImageNet pre-trained image classification model ~\cite{chen2021remote, li2022transunetcd, song2022mstdsnet, shi2020change, ke2022hybrid}. However, the performance of these transformer-based CD methods drastically reduce when directly training from scratch on the target CD dataset. This is likely due to the dense self-attention operation, utilized in these approaches, that has quadratic complexity with respect to tokens, requires longer to converge and prone to over-fitting.  In this work, we look into the problem of designing a transformer-based CD approach that is capable of achieving high performance when trained from \textit{scratch}. 

Most existing transformer-based CD approaches employ a two-stream architecture, where features from both streams are combined through simple operations such as, difference, summation and concatenation~\cite{guo2018learning, bandara2022transformer}. However, these approaches do not employ any explicit feature re-weighting  between both streams. We argue that such naive feature fusion strategies likely struggle to effectively aggregate semantic changes from each stream. In this work, we set out to address the above issues collectively in a single transformer-based CD architecture.

\noindent \textbf{Contributions:}  We propose a transformer-based Siamese two-stream CD framework, named ScratchFormer, that is based on a novel shuffled sparse attention (SSA) operation that strives to better attend to sparse informative regions relevant to the CD task. The proposed SSA performs token-mixing over a sparse subset of shuffled features obtained through a data-dependent feature sampling, enabling optimal CD performance when being trained from scratch directly on the target CD dataset. Furthermore, we introduce a change-enhanced feature fusion module (CEFF) that performs feature fusion based on per-channel re-calibration to enhance the features relevant to the semantic changes, while suppressing the noisy ones.



We perform extensive experiments on four public CD datasets: LEVIR-CD~\cite{chen2020spatial}, DSIFN-CD~\cite{zhang2020deeply},  WHU-CD~\cite{WHUdataset},  and CDD-CD~\cite{Lebedev2018CHANGEDI}. 
Our proposed ScratchFormer approach achieves superior performance over the baseline, highlighting the effectiveness of the proposed contributions. Compared to the baseline, our ScratchFormer achieves an absolute gain of 14.27\% in terms of intersection-over-union (IoU) on the DSIFN-CD dataset. Furthermore, ScratchFormer sets a new state-of-the-art performance on all four datasets. On the DSIFN-CD, our ScratchFormer achieves an IoU score of 90.75\%, outperforming the best existing method~\cite{bandara2022transformer} published in literature by 14.27\%.
Fig.~\ref{fig:baseline_comparison} shows a qualitative comparison between the recent ChangeFormer~\cite{bandara2022transformer} and our ScratchFormer on examples from DSIFN-CD dataset.

\section{Preliminaries}
\noindent\textbf{Problem Formulation:}
Given $I_{pre}$, $I_{post}$ $\in \mathbb{R}^{3\times H\times W}$ as a pair of co-registered satellite images acquired at distinct times $T_1$ and $T_2$, the objective in change detection (CD) is to detect relevant semantic changes between  $I_{pre}$ and $I_{post}$ while ignoring irrelevant changes. Here, the relevant changes include changes in man-made facilities such as, buildings and other constructions. On the other hand, the irrelevant changes include seasonal variations, illumination changes, building shadows, and  atmospheric variations.
Consequently, the goal in CD is to predict a binary mask $\bm{M}$ $\in \mathbb{R}^{H\times W}$ that depicts the semantic  (structural)  changes  between $I_{pre}$ and $I_{post}$. 

\subsection{Baseline Change Detection Framework}\label{sec_baseline}

We adapt the recently introduced transformer-based approach~\cite{bandara2022transformer} as our base framework since it achieves promising performance for the CD task. The base CD framework takes an image pair as input and computes the semantic difference between them using a transformer-based Siamese network. It comprises a transformer encoder, difference feature fusion module, and a decoder. The encoder consists of series of attention layers with each layer comprising the standard self-attention \cite{vaswani2017attention} followed by a feed-forward network. The encoder weights are shared and utilized for computing multi-scale features in both streams (pre-change and post-change). For each scale $i$, the resulting features $F^i_{pre}, F^i_{post}$ from both the streams are input to a difference feature fusion module, which encodes the semantic changes occurring between the streams in the corresponding scale. The difference feature fusion module comprises a feature concatenation followed by two convolutions with batch normalization and ReLU layers in between. It then outputs the feature $F^i_{\text{diff}}$ for scale $i$. These multi-scale features $F^i_{\text{diff}}$ are then input to the decoder, where they are fused  through series of convolution and transpose convolution layers for increasing the spatial resolution of feature maps. Finally, the resulting upsampled features are passed to a mask prediction layer to obtain final semantic binary change map $\bm{M}$.
 \begin{figure}[t!]
\begin{center}
    \includegraphics[width=1\linewidth, keepaspectratio]{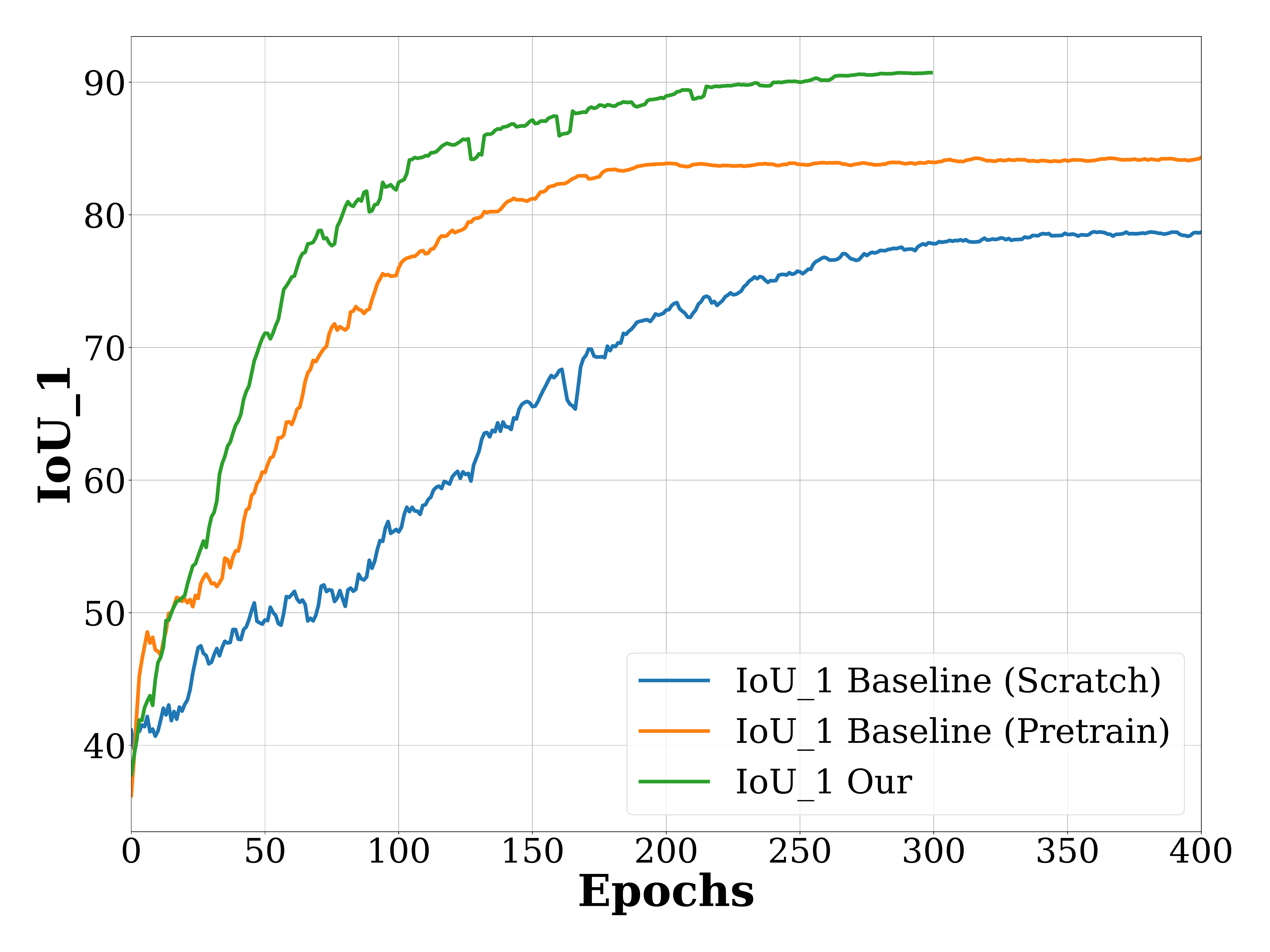}
    \label{fig:iou_plot}
\end{center}
\vspace{-10mm}
\caption{Comparison, in terms of intersection-over-union (IoU) vs. the training epochs, among the baseline trained from scratch, baseline pre-trained$\dagger$ first on another CD dataset and then fine-tuned, and our approach on the DSIFN-CD. Compared to the baseline employing pre-training, training the baseline from scratch results in inferior convergence in terms of CD performance. Our approach despite being trained from scratch achieves superior convergence in terms of CD performance compared to both variants of baseline. For instance, with only 10\% of the training time our approach achieves similar CD performance to that of the final results obtained from the baseline trained from scratch. 
}\label{fig:intro_fig}
\vspace{-1mm}
\end{figure}

\begin{figure*}[t!]
  \centering
    \includegraphics[width=\textwidth]{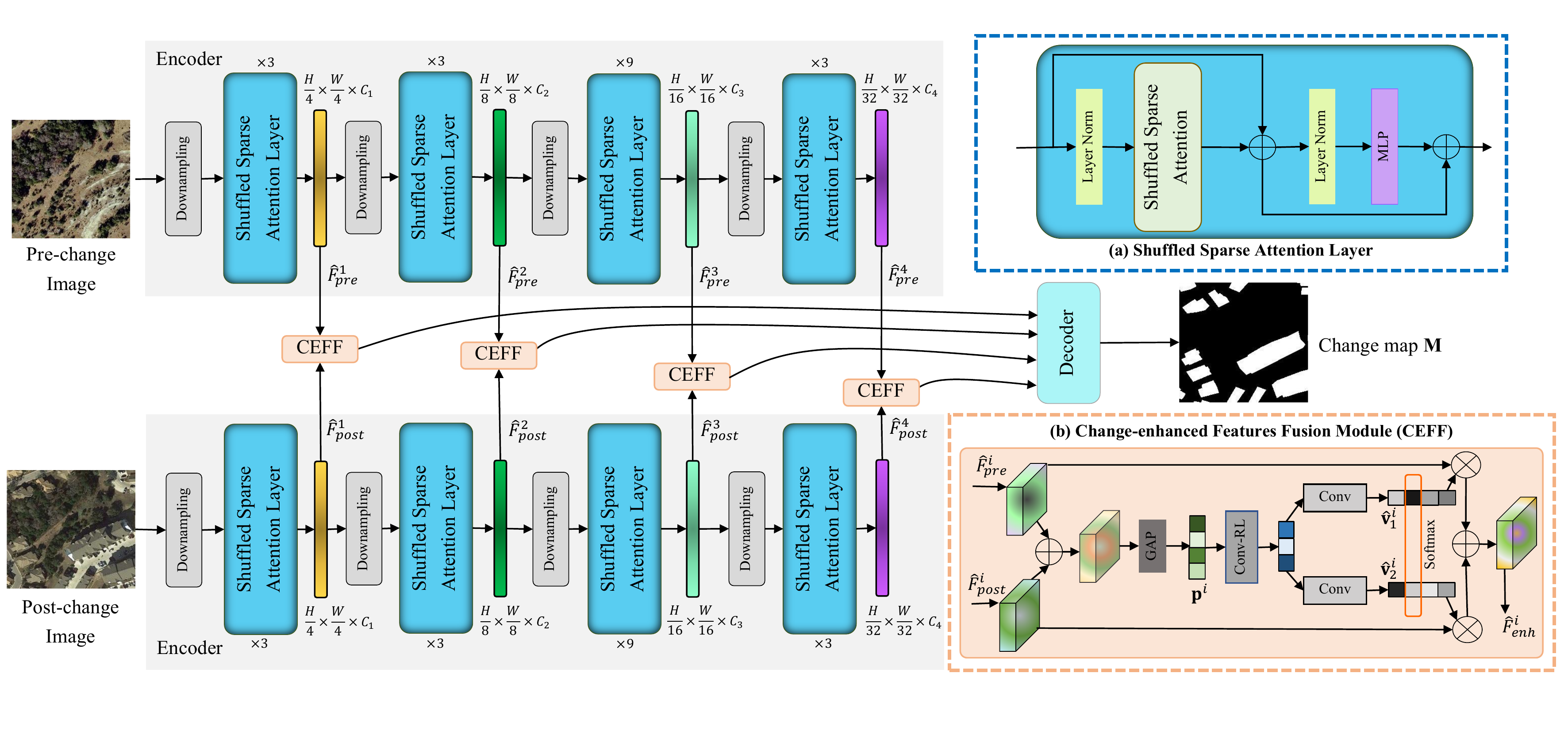} 
  \caption{Overall architecture of our \textbf{ScratchFormer} framework for Change Detection. Our ScratchFormer takes two inputs, pre- and post-change images, and predicts a binary semantic change map for the corresponding image pair. ScratchFormer consists of a Siamese-based hierarchical encoder having four different stages, a change-enhanced feature fusion (CEFF) module, and a decoder for predicting binary change map. 
  The focus of our design is the introduction of a \textit{shuffled sparse attention} (SSA) layer (Sec. \ref{sec:structure_sparse_attention}) in the encoder and a \textit{change-enhanced feature fusion} (CEFF) module (Sec. \ref{sec:change_enhanced_feature_fusion}). 
  The  SSA layer comprises shuffled sparse attention (SSA) and a MLP, as shown in (a). SSA performs token-mixing over a sparse data-dependent subset of features at each stage. 
 Our ScratchFormer approach computes SSA features from the two streams  $\hat F^{i}_{pre}$ and $\hat F^{i}_{post}$ at different scales $i$. The outputs of these stages are fused utilizing the CEFF module, as shown in (b). The CEFF module enhances the semantic changes between the features of the two streams by performing a per-channel re-weighting at each scale and outputs enhanced features $\hat F^i_{enh}$. These enhanced features are then input to the decoder for predicting  the final semantic binary change map $\bm M$.}
  \label{fig:proposed_framework}
\end{figure*}

\noindent\textbf{Limitations:} 
As discussed above, the base framework employs  the transformer encoder with the standard self-attention mechanism to capture long-range dependencies in an image. Here, we argue that the standard self-attention mechanism is sub-optimal for the CD task mainly due to following reason. It operates on uniformly sampled dense patches, thereby requiring large training data for optimal convergence in terms of CD performance (see Fig. \ref{fig:intro_fig}).  
The recent ChangeFormer~\cite{bandara2022transformer}  alleviate this issue by performing pre-training on one (source) CD dataset followed by fine-tuning on another (target) CD data. However, this increases the training time when including the cost of pre-training on another CD dataset as well. Furthermore, despite being trained from scratch the proposed ScratchFormer outperforms our baseline accuracy that is achieved through a pre-training step on another CD dataset, with less than 50\% of the training time.

\section{Method}
\subsection{Motivation}
To motivate our proposed approach, we distinguish two properties especially desired when designing a transformer-based change detection (CD) method.

\noindent\textbf{Rethinking Attention for CD Task:}  As discussed earlier, the conventional self-attention may lead to 
sub-optimal performance when training from scratch directly on the target CD dataset, likely due to difficulty in capturing the inherent inductive biases in the small CD dataset. Moreover, the standard self-attention typically operates on uniformly sampled dense patches that may have difficulties to learn a rich feature representation encoding diverse shape objects with inconsistent appearance in remote sensing scenes having sparse informative regions. Therefore, rethinking the design of the self-attention is desired to effectively learn a rich feature representation by attending to sparse informative regions in remote sensing CD images.
\noindent\textbf{Semantic Change-enhanced Feature Fusion:} While the above requisite focuses on designing a mechanism to attend the  sparse informative regions for the CD task, the second desirable characteristic aims at capturing the semantic differences between image pairs while ignoring the irrelevant noisy changes. 
To this end, a change-enhanced feature fusion module that explicitly models per-channel inter-dependencies between pre- and post-change images is expected to better ignore the noisy changes while retaining the relevant ones. Next, we present our proposed transformer-based ScratchFormer framework.

\subsection{Overall Architecture}
Fig.~\ref{fig:proposed_framework} shows the overall architecture of our ScratchFormer. The proposed ScratchFormer 
takes  pre- and post-change image pairs ($I_{pre}$,  $I_{post}$) as input. 
It comprises a Siamese-based encoder, a change-enhanced feature fusion (CEFF) module, and a decoder for predicting the binary change map $\bm{M}$. The encoder computes the features at four  stages with different spatial resolutions. At each stage, the features are first spatially downsampled through convolutional layers and then input to the shuffled sparse attention (SSA) layers.     
The ScratchFormer consists of two parallel identical encoder streams with shared weights to generate pre- and post- change features $\hat F^{i}_{pre}$,  $\hat F^{i}_{post}$, respectively at the $i$-th stage of our multi-stage network. The focus of our design is the introduction of a novel \textit{shuffled sparse attention} (SSA) layer  in the encoder  to  perform the self-attention on the data-dependent subset of features to effectively capture the semantic changes for CD task. Furthermore, we propose a \textit{change-enhanced feature fusion} (CEFF) module that re-calibrates the per-channel features of the same scale from both streams ($\hat F^{i}_{pre}$ and $\hat F^{i}_{post}$) and performs enhanced feature fusion to better ignore the noisy changes while retaining the relevant ones.



Our SSA {layer} comprises a shuffled sparse attention (SSA) and a multi-layer-perceptron (MLP), as shown in Fig.~\ref{fig:proposed_framework}-(a).  SSA first performs a data-dependent sampling of features to obtain a subset and then performs token-mixing over the selected subset.
SSA strives to focus on the  sparse informative regions for change detection to achieve optimal convergence with respect to CD performance without requiring pre-training on another CD data. 
The CEFF modules aims to enhance the semantic changes between pre- and post- change features at each stage of the encoder, while suppressing the noisy changes.  The resulting enhanced features from  the CEFF module are re-sized to a common spatial resolution and  passed to the decoder. The decoder has a series of convolution, transpose convolution, and upsampling layers to increase the spatial resolution of the feature maps. Consequently, these upsampled features are  passed to a mask prediction layer to obtain the final  binary  mask $\bm M$. Next, we present our SSA layer. 

\subsection{Shuffled Sparse Attention Layer} \label{sec:structure_sparse_attention}
We  introduce a shuffled sparse attention (SSA) layer within our encoder to capture semantic changes between the input image pairs $I_{pre}$  and $I_{post}$.

As shown in Fig.~\ref{fig:proposed_framework}-(a), it comprises a shuffled sparse attention (SSA) to perform token-mixing,  a multi-layer perceptrons (MLP), and  layer normalization layers. Our SSA performs token-mixing over a sparse subset of  features which are selected based on a data-dependent sampling strategy.  Let,  ${F}^i \in \mathcal{R}^{{H}^i \times {W}^i \times {C}^i}$ be the encoder feature at stage $i$ input to SSA.  Then, our  SSA is computed in two steps. First, we perform a data-dependant sparse sub-sampling of input features with a sparsity factor of $\gamma$ to obtain  feature sub-sets $\Bar{F}^i_{kl}$. Then, we separately perform  self-attention over these  $\gamma^2$ feature subsets $\Bar{F}^i_{kl}$, where $k= \{0, ..., \gamma-1\}$ and $l= \{0,..., \gamma-1\}$. The data-dependant sparse spatial sub-sampling of features is performed as follows:  
\begin{equation}\vspace{-0.2cm}
\begin{aligned}[b]
    \Bar{F}^i_{kl}(\bar x , \bar y ) =  F^i(\gamma \bar x + k+ \Delta x,\; \gamma \bar y +l+ \Delta y )   \\
     \forall \bar x=\{0, ..., \frac{H^i} {\gamma}-1 \}\:\: \text{and} \:\:  \forall \bar y=\{0, ..., \frac{W^i} {\gamma}-1 \}
   \end{aligned}
\end{equation}
Here, ($\Delta x$, $\Delta y$), represents the data-dependent position  offsets which  
are predicted using learnable parameters $\theta_{\text{\textit{offset}}}$ as $\Delta z = \theta_{\text{\textit{offset}}}(F^i)$. The predicted offsets $\Delta z \in \mathcal{R}^{{H}^i \times {W}^i \times {2}}$ have two channels depicting the horizontal and vertical position offsets  at each pixel, which are clipped to limit the maximum distance from the current feature  location. Then, the position offsets $\Delta x$, $\Delta y$ are obtained as: \begin{equation*}
\begin{aligned}
\Delta x = \Delta z(\gamma \bar x + k, \gamma \bar y + l, 1)\\
    \Delta y = \Delta z(\gamma \bar x + k, \gamma \bar y + l, 2)\\
   \end{aligned}
\end{equation*}


The resulting sparse-sampled features $\Bar{F}^i_{kl}$ are   used to compute self-attention \cite{vaswani2017attention} ($Attention (.)$) over the $\gamma^2$ sparse windows as follows:
\begin{equation}
     \hat F^i_{kl} = Attention ( \Bar{F}^i_{kl})
\end{equation}
These attended features $\hat F^i_{kl}$ from $\gamma^2$ feature sub-sets are then shuffled back to full resolution feature map  to obtain $\hat{F}^i\in \mathcal{R}^{{H}^i \times {W}^i \times {C}^i}$. Here, the data-dependent position offsets aids to adaptively sample dense features from regions likely having semantic changes, whereas the  sparse sampling helps to  efficiently maintain the global receptive field.  
 Due to the sparse sampling, we perform $\gamma^2$ self-attention operations and in each self-attention operation the number of tokens are reduced by a factor of $\gamma^2$, leading to a $O(\gamma^2)$ reduction in the overall computation. Our SSA enables faster convergence due to its sparse structure allowing self attention to focus on the sub-sampled relevant  features. Our proposed ScratchFormer approach employs SSA  layers at each stage of the encoder and computes pre- and post- change features $\hat{F}^{i}_{pre}$,  $\hat{F}^{i}_{post}$, from both streams of the encoder. These features are then fused by the change-enhanced feature fusion module  described next. 

\subsection{Change Enhanced Features Fusion Module} \label{sec:change_enhanced_feature_fusion}
As discussed earlier, given the diverse nature of the changes in real-world scenarios that can possibly occur in the image pairs, detecting high-level semantic changes while ignoring the noisy  ones is one of the major challenges in the CD task. Therefore, it is desired to effectively fuse the features from pre- and post-change feature streams of the encoder. Within several existing transformers-based CD methods~\cite{chen2021remote, bandara2022transformer, yan2022fully}, multi-level feature fusion between  pre- and post change features is performed through difference, summation or concatenation operations. Similarly, the base framework also introduces a difference module employing  concatenation across channel dimension for the feature fusion. We argue that such a fusion of the features from both streams without explicitly re-weighting the channels from each stage is sub-optimal for the CD task. To this end, we introduce a change-enhanced feature fusion module (CEFF) that performs per-channel re-weighting to enhance the channels  having higher semantic changes, while suppressing the channels capturing noisy changes.   

Fig~\ref{fig:proposed_framework}-(b) shows the structure of our change-enhanced feature fusion module (CEFF). The CEFF module is introduced at all four stages of the encoder to fuse the features at each stage. 
In our CEFF module,  we first  combine the pre-  and post- change features  $\hat{F}^{i}_{pre}$,  $\hat{F}^{i}_{post}$ through addition,  and then perform  global average pooling ($GAP$) to obtain a global feature vector  ${\textbf{p}}^i$ as follows:
\begin{equation}
    {\textbf{p}}^i = GAP(\hat{F}^i_{pre} + \hat{F}^i_{post}),
\end{equation}
We input  ${\textbf{p}}^i$ feature vector to  shared Conv-ReLU layers to reduce the number of channels. Afterwards, these reduced features are passed to  separate $1\times1$ conv  layers to  obtain the channel weights for both streams $\textbf{v}^i_1$, $\textbf{v}^i_2$  as follows: 
\begin{equation}
\begin{aligned}
    {\Bar{\textbf{p}}}^i = \varphi(\omega_1({\textbf{p}}^i)), \\
   \textbf{ v}^i_1 = \omega_2({\Bar{\textbf{p}}}^i),\;  \textbf{v}^i_2 = \omega_3({\Bar{\textbf{p}}}^i),
\end{aligned}
\end{equation}
where, $\omega_1$, $\omega_2$, and $\omega_3$ are the convolutional weights, and $\varphi$ represents the ReLU activation function. Here,  $\textbf{v}^i_1 \in \mathcal{R}^{C^i\times 1}$, and  $\textbf{v}^i_2 \in  \mathcal{R}^{C^i\times 1}$ refers to the un-normalized channels re-weighting factors predicted for the pre- and post-change features at stage $i$. These un-normalized weights are then normalized by per-channel softmax across both streams.  
i.e,  
\begin{equation}
\begin{aligned}
    \hat{\textbf{v}}^i_1(j) =\frac{\text{exp}(\textbf{v}^i_1(j))}{\text{exp}(\textbf{v}^i_1(j)) + \text{exp}(\textbf{v}^i_2(j))}\\
     \hat{\textbf{v}}^i_2(j) =\frac{\text{exp}(\textbf{v}^i_2(j))}{\text{exp}(\textbf{v}^i_1(j)) + \text{exp}(\textbf{v}^i_2(j))}\\
     \forall j=\{1, ..., C^i\}
   \end{aligned}
\end{equation}
where, $j$ is the channel index and $\text{exp}$ denotes the exponential function. 
These normalized weights $\hat{\textbf{v}}^i_1 \in \mathcal{R}^{C^i\times 1}$ and $\hat{\textbf{v}}^i_2 \in \mathcal{R}^{C^i\times 1}$ are used to perform channel re-weighting of   $\hat{F}^i_{pre}$ and $\hat{F}^i_{post}$ followed by feature fusion through addition to generate the enhanced features  $\hat{F}^i_{enh}$ as:
\begin{equation}
   \hat F^i_{enh} =\hat{\textbf{v}}^i_1 \hat F^i_{pre} + \hat{\textbf{v}}^i_2 \hat F^i_{post},
\end{equation}

The resulting enhanced features from the CEFF module at all stages are then resized to a fixed spatial resolution and passed to the decoder that performs feature upsampling  and change map prediction.  
\begin{table*}[t!]
\centering
\caption{State-of-the-art comparison on DSIFN-CD, LEVIR-CD, WHU-CD, and CDD-CD datasets. For each dataset, we report the results in terms of F1, IoU and OA metrics. Our ScratchFormer approach that is trained from scratch and does not require any pre-training performs favorably against existing methods and achieves state-of-the-art performance. The best two results are in red and blue, respectively.}
\label{tbl:comaprison_on_LEVIR_DSIFN}
\scalebox{0.9}{
\begin{tabular}{|l|ccc|ccc|ccc|ccc|} \hline
\multicolumn{1}{|l|}{\multirow{2}{*}{Method}} & \multicolumn{3}{c|}{DSIFN-CD} & \multicolumn{3}{c|}{LEVIR-CD} & \multicolumn{3}{c|}{WHU-CD}  & \multicolumn{3}{c|}{CDD-CD}    \\  \cline{2-13} 
\multicolumn{1}{|l|}{} & F1  & OA  & IoU & F1  & OA & IoU & F1  & OA & IoU  & F1   & OA & IoU  \\  \cline{1-1} \hline \hline 
{TransUNetCD \cite{li2022transunetcd} }  & 66.62 & - &  57.95  & \textcolor{blue}{91.11}  & - & \textcolor{blue}{83.67} & \textcolor{red}{93.59}  &   -& \textcolor{blue}{84.42} &   \textcolor{blue}{97.17}    &  - &  \textcolor{blue}{94.50}\\
DASNet    \cite{chen2020dasnet}    &  - & -& -  &  - &  - &  - & -  &-& - & 92.70   &  \textcolor{blue}{98.2}  & -    \\
H-TransCD  \cite{ke2022hybrid}  & - & -& -& 90.06 & 99.00 & 81.92 & -&  - &    - &-&-& - \\
STANet   \cite{chen2020spatial}  & -& - & - & 87.3 & - & - & - &  -&  -&   - &  -  &   -   \\
SNUNet   \cite{fang2021snunet}   & -&-&-&- & -&-& - &  -&   - & 83.4 & -&      -   \\
BIT  \cite{chen2021remote}    &  69.26   & 89.41  & 52.97  & 89.31  & 98.92 & 80.68  & 83.98  & \textcolor{blue}{98.75}  & 72.39  & -  & - & -   \\ 
IFNet \cite{zhang2020deeply} & 67.33 & {88.86} & - & - & - & - & - &-  & -& 90.30 & 97.71 & -  \\
MSTDSNet   \cite{song2022mstdsnet}  & - & - & - & 88.10  & - & 78.73 & - & - & -& - &   -&    - \\
ChangeFormer  \cite{bandara2022transformer}   & \textcolor{blue}{86.67}  & \textcolor{blue}{95.56} &  \textcolor{blue}{76.48} & 90.40  & \textcolor{blue}{99.04} & 82.48   & -  & -  & - & - & - & -    \\
\hline
\textbf{ScratchFormer (ours)} &  \textcolor{red}{95.15} & \textcolor{red}{98.36}  &  \textcolor{red}{90.75} & \textcolor{red}{91.78} &  \textcolor{red}{99.17} &  \textcolor{red}{84.81}  &  \textcolor{blue}{92.16} &  \textcolor{red}{99.40}  &  \textcolor{red}{85.46}  &  \textcolor{red}{98.14} &  \textcolor{red}{99.56} &  \textcolor{red}{96.34} \\  \hline
\end{tabular}}
\end{table*}


\section{Experiments}
\subsection{Experimental Setup}
\noindent\textbf{Datasets:}
The large-scale \noindent\textit{LEVIR-CD \cite{chen2020spatial}:} dataset is for building change detection. It contains 637 high resolution (0.5m per pixel) image pairs taken from Google Earth with size of 1024x1024. In our experiments, we use the   non-overlapping cropped patches of 256x256, having default data split of train, validation, and test equal to 7120, 1024, and 2048, respectively. The \noindent\textit{DSIFN-CD \cite{zhang2020deeply}:} dataset
is for binary change detection and contains six high-resolution (2m) satellite image pairs from six cities of China. We used the cropped version of the dataset having image size of 256x256 resulting in train, validation and test set of size 14400, 1360, and 192 image pairs, respectively. The \noindent\textit{CDD-CD  \cite{Lebedev2018CHANGEDI}:} dataset comprises 11 seasonal varying image pairs including, 7 image pairs of size  4725x2700 pixels and 4 image pairs of size 1900x1000. The image pairs are clipped into 256x256 with data split of 10000, 3000, and 3000 for train, validation, and test set, respectively. The \noindent\textit{WHU-CD \cite{WHUdataset}:} dataset is for building-related change detection and consists of one high-resolution (0.075m) image pair of size 32507x15354 pixels. This aerial dataset contains a variety of building architectures of different sizes and colors. The dataset is also available with image pairs of size 256x256 pixels having non-overlapping regions and data split of 5947, 743, and 744 image pairs for train, validation and test set, respectively.

\noindent\textbf{Evaluation Protocol:}
Following \cite{bandara2022transformer}, we evaluate change detection (CD) results in terms of \textit{change class} F1-score, \textit{change class} Intersection over Union (IoU) and overall accuracy (OA) on all the datasets. Among these evaluation metrics, the \textit{change class} IoU is the most challenging metric for the CD task. 


\noindent\textbf{Implementation Details:}
Our ScratchFormer takes a pair of images of size $256 \times 256 \times 3$, resizes the input images to $512 \times 512 \times 3$ and computes the features for the two streams at  four stages (having 3, 3, 9, and 3 SSA layers), which outputs the features with 64, 128, 320, and 512 channels, respectively. In the proposed SSA, the  sparsity factor is calculated as $\gamma = 2^n$, where $n>0$.
The decoder outputs the prediction maps with two channels having a spatial resolution of $512 \times 512$ which is resized to $256 \times 256$ to compute the metric scores. The model is trained using pixel-wise cross-entropy loss function. During training, we employ standard data augmentations including, random scale crop, Gaussian blur, random flip, random re-scale, and random color jitter.
We train our network using random initialization on 4 NVIDIA A100 GPUs.
Following~\cite{bandara2022transformer}, we use the AdamW optimizer with a weight decay 0.01 and beta values equal to (0.9, 0.999).   We set the batch size 16, initial learning rate to 4.1e-4, and train for 300 epochs.  
In our experiments, we used linear decay to decrease the learning rate till the last epoch. 
At the inference stage, we freeze the model weights, resize the input image pair of sizes $256 \times 256 \times 3$ to $512 \times 512 \times 3$ and pass 
 to our ScratchFormer model to get the $512 \times 512 \times 2$ prediction maps which are resized back to original spatial resolution of $256 \times 256$. The binary change mask $\bm M$ is computed using a pixel-wise argmax operation along the channel dimension. A well-documented code along with trained models will be publicly released.

\begin{figure*}[t!]
  \centering
    \includegraphics[width=1\textwidth]{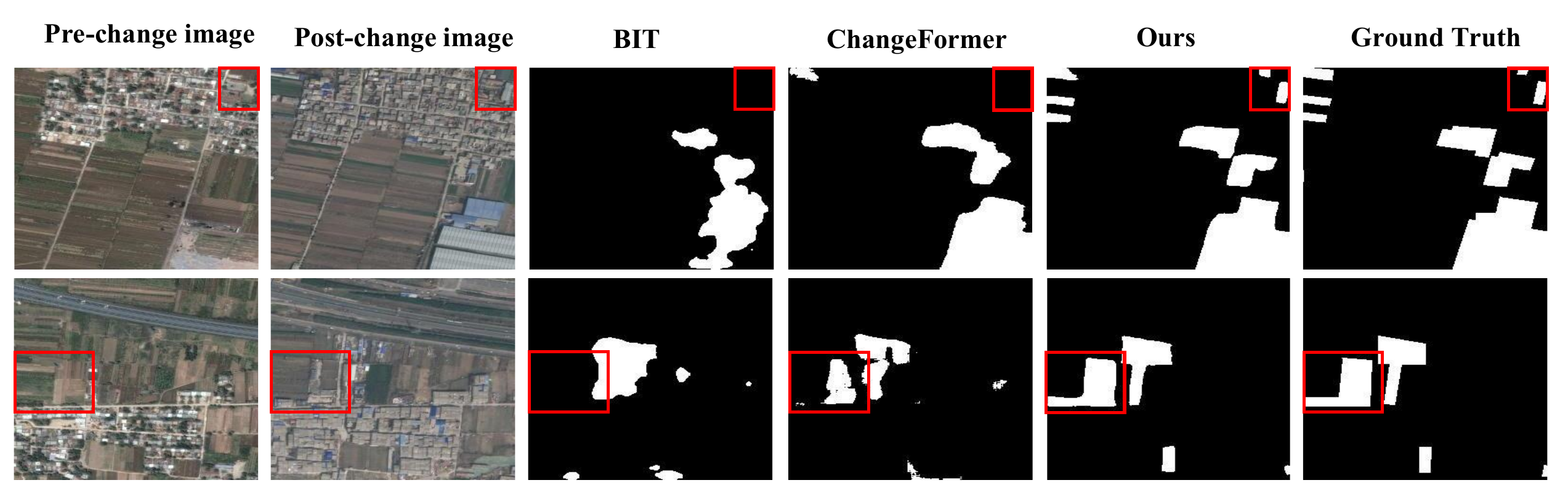}
  \caption{Qualitative comparison on DSFIN-CD. we compare our ScratchFormer with   BIT \cite{chen2021remote} and ChangeFormer \cite{bandara2022transformer}.  We observe ScratchFormer to better detect the semantic changes (marked in red box) with clear boundaries between the pre- and post-change images, compared to other methods. Additional results are presented in the supplementary.
  } 
  \label{fig:comparison_with_SOTA_on_dsifn}
\end{figure*}
\begin{figure*}[t!]
  \centering
    \includegraphics[width=1\textwidth]{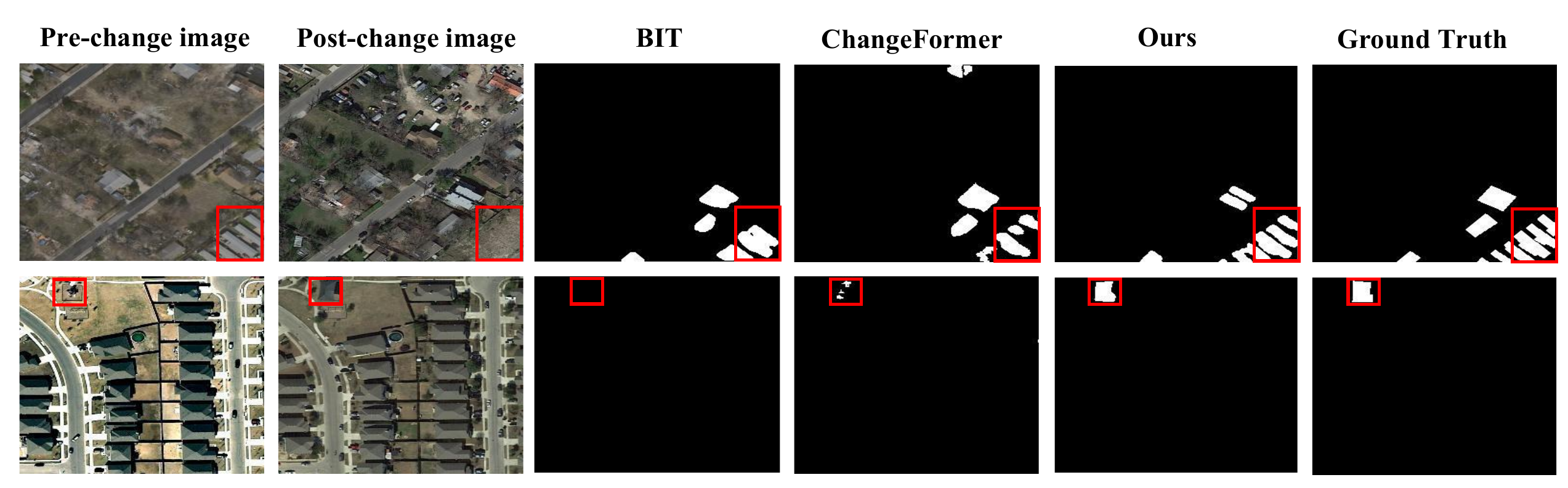} 
  \caption{Qualitative comparison on LEVIR-CD. We compare our ScratchFormer with BIT \cite{chen2021remote} and ChangeFormer \cite{bandara2022transformer}. Our ScratchFormer provides improved CD performance by accurately detecting the relevant changes (marked in red box) with clear boundaries, compared to existing methods. Additional results are presented in the supplementary.
  } 
  \label{fig:comparison_with_SOTA_on_levir}
\end{figure*}

\subsection{State-of-the-art Comparison}

\noindent\textbf{Comparison on DSIFN-CD:}
We compare our approach with both CNN-based and  transformer-based state-of-the-art methods over DSIFN-CD. Tab.~\ref{tbl:comaprison_on_LEVIR_DSIFN} presents the results. The CNN-based IFNet \cite{zhang2020deeply} obtains F1 score of 67.33\%. We observe that  recent transformer-based methods achieve better IoU performance. For instance, BIT \cite{chen2021remote}, TransUNetCD \cite{li2022transunetcd}, and ChangeFormer \cite{bandara2022transformer} achieve IoU scores of 52.97\%, 57.95\%, and 76.48\%, respectively. Our ScratchFormer outperforms these recent methods and achieves consistent improvements in terms of all metrics.  Notably, our ScratchFormer achieves absolute gains of  8.48\%, 2.8\%, and 14.27\% in terms of F1,  OA, and IoU compared to ChangeFormer \cite{bandara2022transformer}. It is worth mentioning that our approach  here is trained from scratch without using any pre-training on another CD dataset. On this dataset, ScratchFormer sets a new state-of-the-art performance with a significant gain obtained in the challenging F1 and IoU metrics. 

\noindent\textbf{Comparison on LEVIR-CD:}
Here, we present the state-of-the-art comparison on the LEVIR-CD dataset (Tab.~\ref{tbl:comaprison_on_LEVIR_DSIFN}). 
Among recent transformer-based CD methods, H-TransCD \cite{ke2022hybrid}, BIT \cite{chen2021remote}, and ChangeFormer \cite{bandara2022transformer},  obtain IoU scores of 81.92\%, 80.68\%, and 82.48\%, respectively. Our ScratchFormer obtains an IoU score of 84.81\% with an absolute gain of 2.33\% over the recently published method in literature  ChangeFormer \cite{bandara2022transformer}.

\noindent\textbf{Comparison on WHU-CD:}Here, we present the state-of-the-art comparison on the WHU-CD dataset (Tab.~\ref{tbl:comaprison_on_LEVIR_DSIFN}). Among existing transformer-based methods, BIT \cite{chen2021remote}  and TransUNetCD \cite{li2022transunetcd} achieve IoU score of 72.39\% and 84.42\%, respectively. 
In comparison, our ScratchFormer that is trained from scratch through random initialization on this dataset achieves favorable performance against existing methods with an IoU score of 85.46\%.


\noindent\textbf{Comparison on CDD-CD:}
Lastly, we report results (Tab.~\ref{tbl:comaprison_on_LEVIR_DSIFN}) on the CDD-CD dataset. Among  CNN-based approaches, DASNet \cite{chen2020dasnet} achieves F1 score of 92.70\%. Among transformer-based methods, TransUNetCD \cite{li2022transunetcd} achieves IoU score of 94.50\%, which achieves this performance by employing improved ResNet50 backbone. In comparison, our ScratchFormer trained from scratch achieves an IoU score of 96.34\%.

\noindent\textbf{Qualitative Comparison:}
Fig.~\ref{fig:comparison_with_SOTA_on_dsifn} and \ref{fig:comparison_with_SOTA_on_levir} present qualitative comparisons of our ScratchFormer  with BIT \cite{chen2021remote}   and ChangeFormer \cite{bandara2022transformer} on DSIFN-CD and LEVIR-CD datasets, respectively. We observe that our ScratchFormer is able to accurately detect  changes occurring at multiple scales in these complex example scenes. For instance, in Fig. \ref{fig:comparison_with_SOTA_on_dsifn} (first row) and Fig. \ref{fig:comparison_with_SOTA_on_levir} (second row), our method better localizes the semantic changed regions, compared to existing methods, which demonstrates the efficacy of our method due to the employment of our novel shuffled sparse attention.

\subsection{Ablation Study}
Here, we present  ablation study to validate the effectiveness of our contributions over  DSIFN-CD and LEVIR-CD dataset. Tab.~\ref{tbl:ablation_dsifn} shows the baseline comparison. The baseline approach (Sec. \ref{sec_baseline}) when trained from scratch using random initialization achieves IoU score of 78.11\%  (row4) over DSIFN-CD dataset. The results of the baseline approach are improved to 83.94\% (row 3) when first pre-training it on LEVIR-CD and then finetuning it on the DSFIN-CD (target) dataset. When integrating our SSA  layer (\ref{sec:structure_sparse_attention}) into the baseline, the results are significantly improved to 86.62\% in terms of IoU score (row 5).
Our final ScratchFormer which includes both contributions (SSA and CEFF) and  trained from scratch leads to a significant improvement in performance by achieving an IoU score of 90.75\%. These results demonstrate the effectiveness of our contributions. Similarly, our method obtains a consistent gain by introducing the proposed SSA and CEFF to the baseline over the LEVIR-CD dataset.  In addition to the baseline comparison, we also report the results of ChangeFormer using both pre-training and training from scratch. Our ScratchFormer achieves consistent gain in performance on all three metrics over the ChangeFormer.


We also conduct an experiment to estimate the optimal sparsity of our SSA by varying the sparsity factor $\gamma$ (2, 4, and 8) as shown in Table \ref{tbl:ablation_grid_size}. We observe setting the value of $\gamma$ to 4 to provide optimal performance for DSIFN and LEVIR datasets. Therefore, we fix the $\gamma$ and use the same value throughout our experiments. 
We further perform an experiment to compare our CEFF module with standard addition, subtraction, and concatenation based techniques. Here, addition, subtraction, and concatenation are performed for $\hat F^i_{pre}$ and $\hat F^i_{post}$, and passed  to two convolutional layers. Tab.~\ref{tbl:effect_of_ceff_dsifn} shows the comparison. Our CEFF that utilizes feature channel re-weighting achieves superior performance compared to these techniques. 

\begin{table}[]
\centering
\caption{Ablation study on the DSIFN and LEVIR datasets. Here, we show the impact of integrating our contributions to the baseline. $\dagger$ denotes that the model is pre-trained first on another CD dataset and then finetuned on the target CD dataset. The integration of our SSA (row 5) into the baseline (row 4) leads to consistent gain in performance. Our final approach ScratchFormer (row 6) that comprises both SSA and CEFF achieves a significant improvement in performance over the baseline. Here, we also report ChangeFormer with and without pre-training. Best two results are in red and blue, respectively.
}
\label{tbl:ablation_dsifn}
\setlength{\tabcolsep}{10pt}
\adjustbox{width=1\columnwidth}{
\begin{tabular}{lcccccc} \toprule[0.15em]
\multicolumn{1}{l}{\multirow{2}{*}{Method}} & \multicolumn{3}{c}{DSIFN-CD} & \multicolumn{3}{c}{LEVIR-CD}   \\  \cline{2-7} 
\multicolumn{1}{l}{} & F1  & OA  & IoU & F1  & OA & IoU  \\  \cline{1-1}  \toprule[0.1em]
ChangeFormer \cite{bandara2022transformer} $\dagger$  &  86.67 & 95.56  & 76.48 & 90.40 & 99.04 & 82.48 \\
ChangeFormer \cite{bandara2022transformer} &  81.24 &  93.54  & 68.41   & 84.97 & 98.52 & 73.86 \\
\midrule
Baseline $\dagger$  & 91.27   & 97.07   & 83.94   &90.84 &99.08 & 83.01 \\ 
Baseline   & 87.71  & 95.79   & 78.11    & 90.68 & 99.07 & 82.95 \\ 
Baseline + SSA (Sec. \ref{sec:structure_sparse_attention})  & \textcolor{blue}{92.83}  & \textcolor{blue}{97.58}   & \textcolor{blue}{86.62}  & \textcolor{blue}{91.18}  & \textcolor{blue}{99.12}   & \textcolor{blue}{83.79}\\
\hline
Baseline + SSA+ CEFF (\textbf{ScratchFormer})    & \textcolor{red}{95.15}  & \textcolor{red}{98.36}   & \textcolor{red}{90.75}   & \textcolor{red}{91.78}  & \textcolor{red}{99.17}   & \textcolor{red}{84.81} \\ \bottomrule[0.15em]
\end{tabular}
}
\end{table}


\begin{table}[]
\centering
\caption{Comparison of the sparsity $\gamma$ over DSIFN and LEVIR datasets. The sparsity $\gamma=4$ achieves superior performance. The best results are in bold.}
\label{tbl:ablation_grid_size}
\setlength{\tabcolsep}{10pt}
\adjustbox{width=1\columnwidth}{
\begin{tabular}{lcccccc} \toprule[0.15em]
\multicolumn{1}{l}{\multirow{2}{*}{$\gamma$}} & \multicolumn{3}{c}{DSIFN-CD} & \multicolumn{3}{c}{LEVIR-CD}   \\  \cline{2-7} 
\multicolumn{1}{l}{} & F1  & OA  & IoU & F1  & OA & IoU  \\  \cline{1-1}  \toprule[0.1em]
$\gamma$=2   & 93.91  & 97.95   & 88.53   & 91.55 & 99.15 & 84.42 \\ 
$\gamma$=4   & \textbf{95.15}  & \textbf{98.36}   & \textbf{90.75}    & \textbf{91.78} & \textbf{99.17} & \textbf{84.81} \\
$\gamma$=8    & 94.34  & 98.10   & 89.29   & 91.68 & 99.16 & 84.64 \\  \bottomrule[0.15em]
\end{tabular}
}
\end{table}



\begin{table}[]
\centering
\caption{ Comparison of CEFF with the subtraction, addition, and concatenation based techniques on DSFIN-CD. CEFF achieves superior performance on all metrics.  Best two results are in red and blue, respectively.}
\label{tbl:effect_of_ceff_dsifn}
\setlength{\tabcolsep}{8pt}
\adjustbox{width=\columnwidth}{
\begin{tabular}{lccc} \toprule[0.15em]
\textbf{Method} & \textbf{F1} & \textbf{OA}  & \textbf{IoU} \\  \toprule[0.1em]
Difference module  with Subtraction & 80.23   & 88.08    & 68.89 \\
Difference module with Addition & 93.00    & 87.34 & 78.53  \\
Difference module  with Concatenation & \textcolor{blue}{92.83}  & \textcolor{blue}{97.58}   & \textcolor{blue}{86.62}   \\
CEFF  & \textcolor{red}{95.15}  & \textcolor{red}{98.36}   & \textcolor{red}{90.75} \\ \bottomrule[0.15em]
\end{tabular}
} 
\end{table}

\section{Relation to Prior Art}
\noindent\textbf{CNN-based Approaches:} 
Convolutional neural networks have attained much popularity in remote sensing change detection due to intrinsic properties to capture discriminative features \cite{shi2020change}. Chen et al. \cite{chen2020dasnet} propose a dual attention mechanism within Siamese CNN to encode long-range dependencies. Fang et al. \cite{fang2021snunet} propose adense Siamese network that uses channel attention to refine  features. A feature pyramid with attention mechanism is proposed to encode long-range dependencies in \cite{jiang2020pga}. 
Liu et al. \cite{liu2021super} use multi-scale convolutional attention features to learn the bitemporal feature differences via adversarial learning.  Hou et al. \cite{hou2017change} employ low rank analysis to benefit from deep features for CD. 
Chen et al. \cite{chen2020spatial} introduce Siamese-based network to capture spatial–temporal dependencies for CD. 
Zhang et al. \cite{zhang2020deeply} propose a deep supervised image fusion network for CD. 

\noindent\textbf{Transformers-based Approaches:}
Recently, transformers \cite{vaswani2017attention} have gain popularity for the CD task. 
Chen et al. \cite{chen2021remote} introduce a bitemporal image transformer (BIT) to model context information.  Li et al. \cite{li2022transunetcd} introduce TransUNetCD, which benefits from both transformers and UNet for CD. 
Song et al. \cite{song2022mstdsnet} propose a multi-scale Swin transformer  that uses refined multi-scale features.
Ke et al. \cite{ke2022hybrid} propose a hybrid transformer to capture global context dependencies at multiple scales.
Bandara et al. \cite{bandara2022transformer} propose a hierarchical Siamese transformer to render multi-scale features. Different to existing approaches, we introduce a SSA to effectively capture the inductive CD bias when training from scratch on any change detection (target) dataset. Further, a CEFF module to perform per-channel re-weighting to enhance the feature channels having higher semantic changes, while suppressing the channels encoding noisy changes. 



\section{Conclusion}

We propose an transformers-based Siamese architecture, named ScratchFormer, for the problem of remote sensing change detection (CD). Our ScratchFormer introduces a shuffled sparse attention (SSA) to effectively capture the  inherent characteristics when training from scratch. We further introduce a change-enhanced feature fusion (CEFF) module to perform per-channel feature weighting to enhance the relevant semantic changes, while suppressing the noisy ones. We validate our approach by conducting extensive experiments on four challenging CD datasets. Our extensive qualitative and quantitative results reveal the benefits of the proposed contributions. The proposed ScratchFormer approach achieves  state-of-the-art CD performance on all four CD datasets. A potential future  direction is to explore the problem of change detection in natural images and medical imaging.



{\small
\bibliographystyle{ieee_fullname}
\bibliography{main_paper}
}

\end{document}